\documentclass[letterpaper]{article}
\usepackage[T1]{fontenc}
\usepackage{newtxtext}
\usepackage{helvet}
\usepackage{courier}
\usepackage[hyphens]{url}
\usepackage{graphicx}
\usepackage{amsmath}
\usepackage{amssymb}
\usepackage{natbib}
\usepackage[hidelinks]{hyperref}
\usepackage{microtype}
\usepackage{pdfpages}
\usepackage{booktabs}
\usepackage[table]{xcolor}
\definecolor{retrievergray}{gray}{0.9}

\newcommand{\equalcontrib}{\textsuperscript{*}}
\newcommand{\corresponding}{\textsuperscript{\textdagger}}

\makeatletter
\def\@affiliations{}
\newcommand{\affiliations}[1]{\def\@affiliations{#1}}
\def\maketitle{%
  \par
  \begingroup
    \twocolumn[\@maketitle]
  \endgroup}
\def\@maketitle{%
  \vbox{%
    \hsize\textwidth
    \linewidth\hsize
    \vskip 0.625in minus 0.125in
    \centering
    {\LARGE\bfseries \@title\par}
    \vskip 0.1in plus 0.5fil minus 0.05in
    {\Large\bfseries \@author\par}
    \vskip .2em plus 0.25fil
    {\normalsize \@affiliations\par}
    \vskip 1em plus 2fil}}
\renewenvironment{abstract}{%
  \centerline{\bfseries Abstract}
  \vspace{0.5ex}
  \setlength{\leftmargini}{10pt}
  \begin{quote}
    \small
}{%
  \par
  \end{quote}
  \vskip 1ex
}
\def\section{\@startsection {section}{1}{\z@}{-2.0ex plus -0.5ex minus -.2ex}{3pt plus 2pt minus 1pt}{\Large\bfseries\centering}}
\def\subsection{\@startsection{subsection}{2}{\z@}{-2.0ex plus -0.5ex minus -.2ex}{3pt plus 2pt minus 1pt}{\large\bfseries\raggedright}}
\def\subsubsection{\@startsection{subparagraph}{3}{\z@}{-6pt plus -2pt minus -1pt}{-1em}{\normalsize\bfseries}}
\renewcommand\paragraph{\@startsection{paragraph}{4}{\z@}{-6pt plus -2pt minus -1pt}{-1em}{\normalsize\bfseries}}
\skip\footins 9pt plus 4pt minus 2pt
\def\footnoterule{\kern-3pt \hrule width 5pc \kern 2.6pt}
\labelwidth\leftmargini\advance\labelwidth-\labelsep \labelsep 5pt
\def\@listi{\leftmargin\leftmargini}
\def\@listii{\leftmargin\leftmarginii
  \labelwidth\leftmarginii\advance\labelwidth-\labelsep
  \topsep 2pt plus 1pt minus 0.5pt
  \parsep 1pt plus 0.5pt minus 0.5pt
  \itemsep \parsep}
\def\@listiii{\leftmargin\leftmarginiii
  \labelwidth\leftmarginiii\advance\labelwidth-\labelsep
  \topsep 1pt plus 0.5pt minus 0.5pt
  \parsep \z@
  \partopsep 0.5pt plus 0pt minus 0.5pt
  \itemsep \topsep}
\belowdisplayskip \abovedisplayskip
\def\normalsize{\@setfontsize\normalsize\@xpt{11}}
\def\small{\@setfontsize\small\@ixpt{10}}
\def\footnotesize{\@setfontsize\footnotesize\@ixpt{10}}
\def\scriptsize{\@setfontsize\scriptsize\@viipt{10}}
\def\tiny{\@setfontsize\tiny\@vipt{7}}
\def\large{\@setfontsize\large\@xipt{12}}
\def\Large{\@setfontsize\Large\@xiipt{14}}
\def\LARGE{\@setfontsize\LARGE\@xivpt{16}}
\def\huge{\@setfontsize\huge\@xviipt{20}}
\def\Huge{\@setfontsize\Huge\@xxpt{23}}
\makeatother
\title{GistPilot: Gist-Trace-Guided Adaptive Memory Access for Personalized LLMs}
\author{%
Zeyu Zhang\textsuperscript{1}\equalcontrib,
Shuai Nie\textsuperscript{2}\equalcontrib,
Xinyuan Zhang\textsuperscript{2},
Chen Jiao\textsuperscript{2},
\\
Jie Ding\textsuperscript{2},
Xinyue Li\textsuperscript{1},
Liuji Chen\textsuperscript{1},
Chenxun Deng\textsuperscript{1},
\\
Jiaming Xu\textsuperscript{2},
Peng Zhou\textsuperscript{2},
Junwen Zhang\textsuperscript{2},
Qiang Liu\textsuperscript{1}\corresponding}
\affiliations{%
\textsuperscript{1}Institute of Automation, Chinese Academy of Sciences\\
\textsuperscript{2}Xiaomi\\
\textsuperscript{*}Equal contribution.\quad
\textsuperscript{\textdagger}Corresponding author.\\
\texttt{zeyu.zhang@nlpr.ia.ac.cn}}
\date{}

\begin{document}

\maketitle

\begin{abstract}
External memory systems give large language models (LLMs) access to long user histories, but keep memory outside the model: before retrieval, the model lacks a holistic view of the user and cannot decide whether retrieval is needed or how to query the history. We introduce \textbf{GistPilot}, a personalized memory framework that couples an internal user trace with adaptive external-memory access. User Gist Trace (\textbf{UGT}) compresses each user's history into a fixed-length latent KV prefix available before retrieval. Conditioned on UGT, the Gist-guided Memory Access Policy (\textbf{GAP}) estimates how much question-relevant memory the trace retains and selects direct answering, specific-query retrieval, coarse-query retrieval, or retrieval with the original query. UGT is learned through history reconstruction and context distillation; GAP is trained with a supervised warm start followed by outcome-driven reinforcement learning using counterfactual rollouts and role-specific advantage assignment. Experiments on LongMemEval and PersonaBench show that GistPilot achieves the best overall answer quality in Judge score over the strongest baseline. These results show that a compact internal gist can provide global user context while enabling effective, query-dependent memory access.
\end{abstract}
\noindent\textbf{Project page.} \url{https://github.com/zzFestinaLente/Gist_Pilot}
\section{Introduction}
External memory is widely used to equip large language models (LLMs) with persistent, personalized memory~\citep{packer2023memgpt,zhong2024memorybank,chhikara2025mem0}. Such systems extract, consolidate, and retrieve information from long interaction histories, enabling answers beyond the current context window~\citep{chhikara2025mem0,wu2025longmemeval,maharana2024locomo,tan2025personabench}. In this retrieval-centered design, memory remains outside the model, while selected records, profiles, or summaries are injected into the context on demand~\citep{lewis2020rag,chhikara2025mem0}. This design eases context-window limitations but separates memory storage from awareness. Before retrieval, the model has no internal view of the user's history and cannot assess what it knows or which memories matter to the question.

\begin{figure}[t]
\centering
\includegraphics[width=\columnwidth]{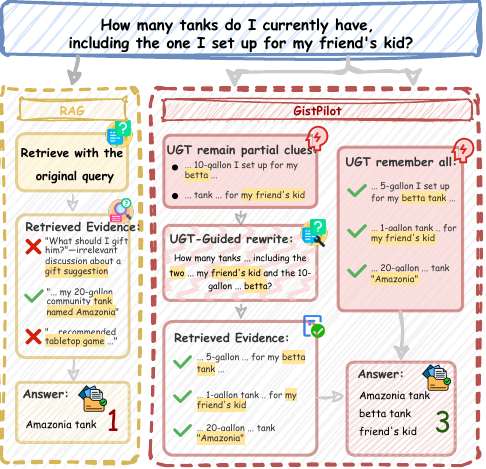}
\caption{Original-query retrieval versus GistPilot.}
\label{fig:case}
\end{figure}

How can a model form an initial awareness of a user's history before explicit retrieval? Human memory offers inspiration. Fuzzy-Trace Theory~\citep{reyna2012newintuitionism} holds that memory is not limited to all-or-nothing exact recall. It distinguishes verbatim traces that preserve details from gist traces that retain underlying meaning and support judgment when details are incomplete. Pattern completion~\citep{horner2015holistic} lets partial cues reactivate related information. Metamemory~\citep{nelson1990metamemory} adds a control layer: monitoring assesses memory accessibility, while control selects the next action accordingly. These judgments directly shape question-answering strategies~\citep{reder1987strategy}.

However, existing personalized memory systems still lack both capabilities. First, direct retrieval and query rewriting remain centered on external memory without a gist-like representation of user history to guide interpretation and activate associations~\citep{lewis2020rag,zhang2026pbr}. Thus, they may retrieve semantically similar but intent-irrelevant records while missing differently expressed memories whose relevance emerges only from user history. Figure~\ref{fig:case} illustrates this failure: the phrase
``my friend's kid'' steers similarity-based retrieval toward an unrelated
gift-suggestion record, while the original query retrieves only the Amazonia
tank and misses the other two tank records needed for a complete answer. Second, many systems follow a fixed path (always retrieving, or rewriting) rather than choosing the best method for one query~\citep{ma2023rewrite,gao2023hyde,jia2024mill,zhang2026pbr}. Rewriting with insufficient cues may fabricate information; original-query retrieval may underuse retained cues; and unconditional retrieval adds cost and noise when internal memory suffices.

To address these limitations, we introduce GistPilot, a personalized memory framework that combines a user gist trace with adaptive external-memory access through two components: User Gist Trace (UGT) and the Gist-guided memory Access Policy (GAP). The UGT compresses each user's long-term history into a fixed-length, user-specific KV memory which provides a holistic user-level signal before retrieval rather than replacing verifiable external-memory records. The GAP estimates question-relevant memory retained in UGT and chooses among four actions: answering directly when memory suffices; generating a precise query from clear cues; producing a broader clue from coarse context; or using the original question without reliable cues, avoiding erroneous or misleading query reformulations. In Figure~\ref{fig:case}, GistPilot answers directly when UGT
retains all relevant tank memory; when only partial cues remain, it uses them to reformulate the query and recover the missing evidence.

We train GistPilot in three stages: UGT construction, supervised warm-start, and outcome-driven reinforcement learning. We first construct a fixed-length UGT for each user through history reconstruction, which writes long-term user information into the latent KV prefix, and context distillation, which makes that information reliably accessible for question answering. We then warm-start GAP with SFT to establish stable memory state prediction and conditional generation behavior before reinforcement learning. Finally, we optimize GAP with DAPO~\citep{yu2025dapo} using counterfactual rollouts over all candidate paths, with rewards for final answer quality and retrieval gains and role-specific advantages for state, query, and answer tokens. This stage learns access decisions and generated content from their downstream utility rather than relying solely on teacher-assigned states.

Our main contributions are as follows:
\begin{itemize}
\item We introduce \textbf{GistPilot} and its \textbf{UGT}, which encodes each user's long-term history as a fixed-length latent KV prefix. UGT provides a coarse-grained, holistic view of the user before external retrieval.
\item We propose the \textbf{GAP}, which estimates how much question-relevant memory UGT retains and accordingly selects different memory access ways based on the guidance of user gist trace.
\item Experiments on LongMemEval and PersonaBench show that GistPilot consistently improves personalized question answering, demonstrating the value of combining user gist trace with adaptive memory access.
\end{itemize}

\section{Related Work}

\paragraph{Personalized Long-Term Memory.}
Long-term conversational memory systems extend beyond the context window by organizing interaction histories in external stores. MemGPT manages tiered memory through OS-inspired control~\citep{packer2023memgpt}, while Mem0 extracts and consolidates salient conversational facts into scalable textual or graph-structured memories~\citep{chhikara2025mem0}. Recent work adds user-specific access signals: PBR uses expression patterns and corpus structure for personalized query expansion~\citep{zhang2026pbr}; and RF-Mem switches between familiarity- and recollection-based retrieval according to retrieval uncertainty~\citep{zhang2026rfmem}. However, personalization remains encoded in explicit text, profiles, or retriever-side scores. GistPilot instead maintains a fixed-length latent user trace available before retrieval to guide external-memory access.

\begin{figure*}[t]
\centering
\includegraphics[width=\textwidth]{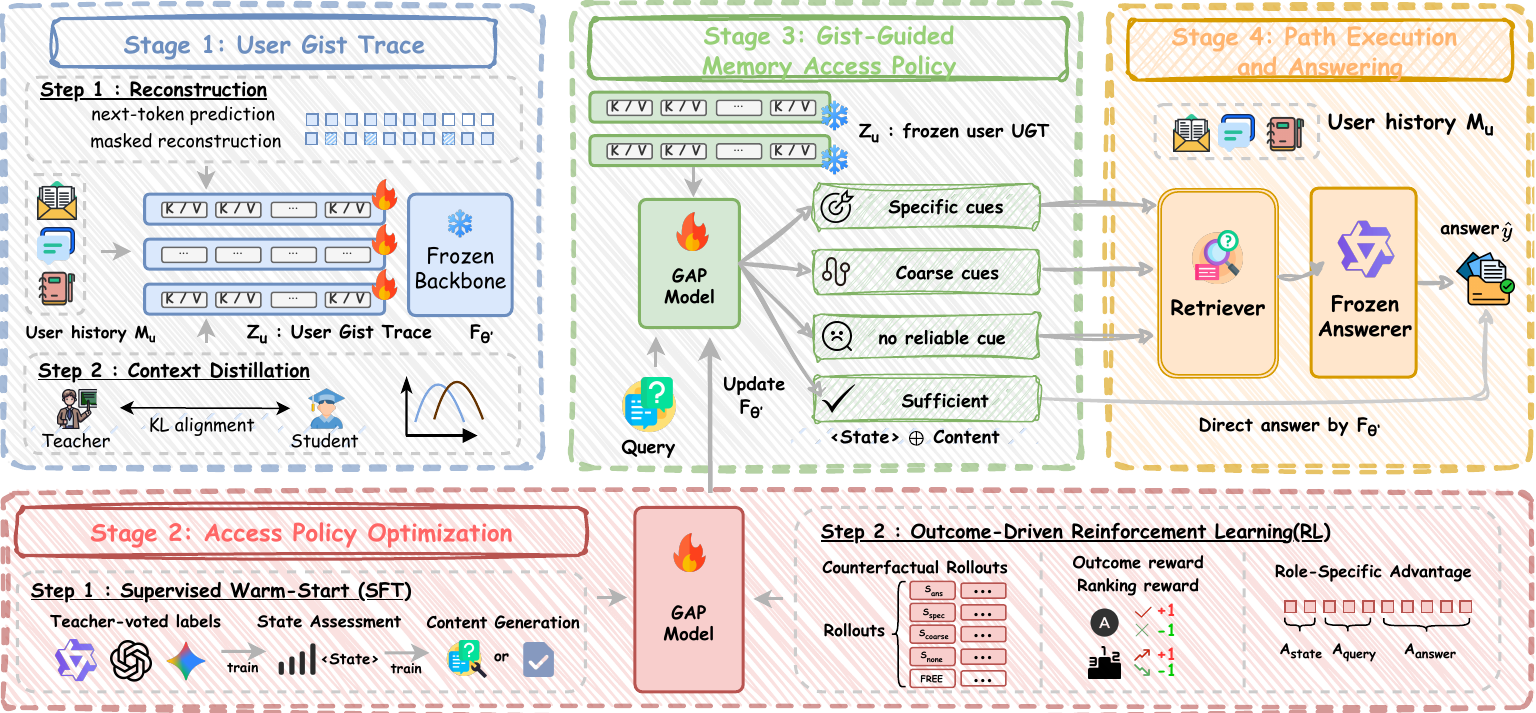}
\caption{Overview of GistPilot, which comprises four stages: constructing the
User Gist Trace (UGT), optimizing GAP through supervised and outcome-driven
learning, selecting an access path from the inferred memory state, and
answering directly or through retrieval.}
\label{fig:overview}
\end{figure*}

\paragraph{Latent Memory and Adaptive Access.}
Continuous memory methods compress long contexts into reusable internal representations. AutoCompressors encode prior context as soft prompts~\citep{chevalier2023autocompressors}, MEMORYLLM uses a fixed-size latent memory pool~\citep{wang2024memoryllm}, and Cartridges distills a corpus into a compact KV cache through self-study and context distillation~\citep{eyuboglu2025cartridges}. These methods reduce repeated processing but treat compressed memory as a uniform context substitute rather than estimate its retained question-relevant memory. Query expansion methods such as HyDE and Query2doc enrich sparse queries with LLM-generated pseudo-documents~\citep{gao2023hyde,wang2023query2doc} but use a fixed strategy. Adaptive RAG methods instead control retrieval through generation uncertainty, self-reflection, or question complexity~\citep{jiang2023flare,asai2024selfrag,jeong2024adaptiverag}. Mem-$\pi$ learns when and what memory guidance to generate from downstream outcomes~\citep{mempi}. However, these decisions rely on generic model knowledge or task difficulty rather than a user-specific memory state. GistPilot couples UGT with GAP, which estimates question-relevant memory retention and selects the most appropriate external memory access strategy under the guidance of UGT.

\section{Method}

\subsection{Problem Setup and Overview}
For a user $u$, let $\mathcal{M}_u=\{m_i\}_{i=1}^{N_u}$ denote the history
accumulated before the current question, where each $m_i$ is a turn, a session,
or any text carrying user-specific information. Given a question $q$ with target
answer $y$, the system produces a prediction $\hat{y}$.

GistPilot answers $q$ through two modules that mirror the two human faculties
identified above, as illustrated in Figure~\ref{fig:overview}. The \textbf{UGT} encodes $\mathcal{M}_u$ into a
fixed-length latent key-value prefix $Z_u$, giving the model a holistic latent trace of the user's history that is available before retrieval. Conditioned on $Z_u$ and $q$, \textbf{GAP} estimates a retention state $s$, which captures how much
question-relevant memory remains in user gist trace, and accordingly produces
the action content $\xi$: a direct answer, a coarse- or fine-grained query
reformulation, or the original query for retrieval. For retrieval-oriented outputs, a frozen retriever $\mathcal{R}$ fetches evidence from
$\mathcal{M}_u$ and a separate frozen answerer $\mathcal{A}$ produces the final
answer.

\subsection{User Gist Trace}

Formally, let the
backbone $\mathcal{F}_{\theta}$ have $L$ Transformer layers. We represent the memory of
user $u$ as a per-layer trainable KV prefix
\begin{equation}
Z_u=\{(K_u^{\ell},V_u^{\ell})\}_{\ell=1}^{L},
\end{equation}
where $K_u^{\ell},V_u^{\ell}$ are keys and values of length $p$ at layer $\ell$.
For an input sequence $X$, layer $\ell$ attends over the concatenation
$K^{\ell}=[K_u^{\ell};K^{\ell}(X)]$ and $V^{\ell}=[V_u^{\ell};V^{\ell}(X)]$, so
every token reads the user memory through ordinary attention. During memory
construction $\mathcal{F}_{\theta}$ stays frozen and only $Z_u$ is optimized.

\paragraph{History reconstruction.}
Over the training sequences $\mathcal{X}_u=\{m_i\}_{i=1}^{N_u}$, we minimize
next-token prediction together with a masked term that discourages reliance on
local surface prefixes:
\begin{equation}
\begin{aligned}
\mathcal{L}_{\mathrm{rec}}(Z_u)
={}&-\sum_{x\in\mathcal{X}_u}\sum_t
\log \mathcal{F}_{\theta}(x_t\mid Z_u,x_{<t}) \\
&+\lambda_{\mathrm{mask}}
\sum_{x\in\mathcal{X}_u}
\mathcal{L}_{\mathrm{mask}}(Z_u;x).
\end{aligned}
\end{equation}
where $\mathcal{L}_{\mathrm{mask}}$ asks the model to restore randomly masked
content. Details are given in \hyperlink{supplement.1}{the supplementary material}.
\paragraph{Context distillation.}
Following the self-study and context distillation methods of
Cartridges~\citep{eyuboglu2025cartridges}, we align the compressed memory with a
teacher through teacher--student distribution matching. We condition $\mathcal{F}_{\theta}$ on each
historical record $m_j$ to construct a synthetic question-answering dataset
$\mathcal{D}_u=\{(m_j,q_j,y_j)\}_{j=1}^{J_u}$. For distillation, the model
conditioned explicitly on $m_j$ serves as the teacher, whereas the model with
access only to $Z_u$ serves as the student. Let $p^{\mathrm{T}}_{j,t}$ and
$p^{\mathrm{S}}_{j,t}$ denote their respective predictive distributions at the
$t$-th answer position:
\begin{equation}
\begin{aligned}
p^{\mathrm{T}}_{j,t}
&=\mathcal{F}_{\theta}(\cdot\mid m_j,q_j,y_{j,<t}),\\
p^{\mathrm{S}}_{j,t}
&=\mathcal{F}_{\theta}(\cdot\mid Z_u,q_j,y_{j,<t}),\\
\mathcal{L}_{\mathrm{distill}}(Z_u)
&=\sum_{j=1}^{J_u}\sum_{t=1}^{|y_j|}
D_{\mathrm{KL}}\!\left(
p^{\mathrm{T}}_{j,t}\,\middle\|\,p^{\mathrm{S}}_{j,t}
\right).
\end{aligned}
\end{equation}
We initialize $Z_u$ by minimizing $\mathcal{L}_{\mathrm{rec}}$, refine it with
$\mathcal{L}_{\mathrm{distill}}$, and then freeze it as the user's persistent
memory for all subsequent stages.

\subsection{Gist-Guided Memory Access Policy}
GAP uses a policy model $\mathcal{F}_{\theta'}$, obtained by full-parameter fine-tuning of the base model, shared across users, to estimate how much information the user gist trace $Z_u$ retains about the current question $q$ and to generate the corresponding action $\xi$. Conditioned on $Z_u$, GAP can move beyond surface-level similarity and formulate queries that retrieve memories relevant to the user's underlying intent.

Concretely, the model emits a structured trajectory that generates the state before the action content:
\begin{equation}
\tau\sim\mathcal{F}_{\theta'}(\cdot\mid Z_u,q),
\qquad
\tau=\langle s\rangle\oplus\xi,
\end{equation}
where $\langle s\rangle$ serializes the retention state $s$, $\xi$ is the action
content generated under that state, and $\oplus$ is concatenation. The action content is determined by the retention state as:
\begin{equation}
\xi=
\begin{cases}
\hat{y}_{\mathrm{dir}}, & s=s_{\mathrm{ans}},\\
\hat{q}_{\mathrm{spec}}, & s=s_{\mathrm{spec}},\\
\hat{q}_{\mathrm{coarse}}, & s=s_{\mathrm{coarse}},\\
q, & s=s_{\mathrm{none}}.
\end{cases}
\end{equation}

Here, $s_{\mathrm{ans}}$ indicates that $Z_u$ retains sufficient
question-relevant memory to answer the question directly, yielding
$\hat{y}_{\mathrm{dir}}$; $s_{\mathrm{spec}}$ indicates that $Z_u$ retains
specific entities, relations, or contextual clues that are insufficient for
answering but sufficient to formulate a targeted retrieval query
$\hat{q}_{\mathrm{spec}}$; $s_{\mathrm{coarse}}$ indicates that only broad
topical or situational cues are recoverable, which are used to construct a
broader query $\hat{q}_{\mathrm{coarse}}$; and $s_{\mathrm{none}}$ indicates
that no reliable query-relevant clue can be recovered from $Z_u$, so the
original question $q$ is used unchanged. The final answer follows the predicted
state:
\begin{equation}
\hat{y}=
\begin{cases}
\hat{y}_{\mathrm{dir}}, & s=s_{\mathrm{ans}},\\[2pt]
\mathcal{A}\big(q,\mathcal{R}(\xi;\mathcal{M}_u)\big),
& s\in\{s_{\mathrm{spec}},s_{\mathrm{coarse}},s_{\mathrm{none}}\},
\end{cases}
\end{equation}
On the retrieval paths, the retriever $\mathcal{R}$ fetches evidence from the user
corpus $\mathcal{M}_u$ using $\xi$, and a separate frozen answerer $\mathcal{A}$
produces the final answer.

\subsection{Access Policy Optimization}
GistPilot optimizes GAP in two stages, a supervised warm-start followed by
outcome-driven reinforcement learning. Throughout training, all user memories
$Z_u$ stay frozen, and only the parameters of the cross-user policy are updated,
fine-tuning the base model $\mathcal{F}_{\theta}$ into the policy model
$\mathcal{F}_{\theta'}$.

\paragraph{Label construction.}
Following the data-generation protocol of LongMemEval~\citep{wu2025longmemeval}, we construct approximately $4{,}000$ training instances and retain only those that pass cross-teacher agreement and quality checks. 

For each instance $(\mathcal{M}_u,Z_u,q_i,y_i)$, the base model $\mathcal{F}_{\theta}$ answers $q_i$ using only $Z_u$, without access to $\mathcal{M}_u$, and verbalizes the relevant memories encoded in $Z_u$. Multiple teacher models then evaluate these outputs against $y_i$ to estimate how much question-relevant memory is retained in the UGT. Then they assign the corresponding memory-state label $s_i^\star$ by majority vote, and generate the corresponding target content $\xi_i^\star$. The prompts, voting procedure, and query-quality filters are detailed in \hyperlink{supplement.1}{the supplementary material}.

\paragraph{Structure-aware warm-start.}
We first warm-start GAP on
$\mathcal{D}_{\mathrm{SFT}}=\{(Z_{u},q_i,s_i^\star,\xi_i^\star)\}_{i=1}^{N}$.
For each sample, the supervised target is the structured trajectory
$\tau_i^\star=\langle s_i^\star\rangle\oplus\xi_i^\star$. We optimize a
token-weighted cross-entropy objective:
\begin{equation}
\mathcal{L}_{\mathrm{SFT}}(\theta')
=
\frac{1}{N_{\mathrm{act}}}
\sum_{i,t}
w_{i,t}\,
\operatorname{CE}\!\left(
\mathcal{F}_{\theta'}\!\left(
\cdot\mid
\substack{
Z_{u},q_i,
\tau_{i,<t}^{\star}
}
\right),
\tau_{i,t}^{\star}
\right).
\end{equation}
where $N_{\mathrm{act}}$ is the total number of unmasked target tokens in the batch and
$w_{i,t}=w_{i,t}^{\mathrm{span}}w_{s_i^\star}^{\mathrm{class}}$.
The span weight controls the supervision assigned to state, query, and answer
tokens, while the class weight associated with the target state $s_i^\star$
compensates for state imbalance. 

Training proceeds
in two stages: state calibration emphasizes the state prefix, and conditioned
generation emphasizes query tokens and, when applicable, direct-answer tokens.
Details are provided in \hyperlink{supplement.1}{the supplementary material}.
\paragraph{Outcome-driven optimization.} Inspired by Mem-$\pi$~\citep{mempi}, we construct counterfactual rollouts by forcing each of the four state prefixes and executing the corresponding answering or retrieval path for the same question. These rollouts estimate the relative utility of all candidate paths. We additionally sample free rollouts under the policy's own state prediction and use the counterfactual outcomes to optimize its state selection and state-conditioned generation, rather than relying solely on teacher-assigned states. Each rollout receives the reward \begin{equation} \widetilde{R}_j= R_{\mathrm{ans},j} +\lambda_{\mathrm{ret}}\, \mathbb{I}[s_j\in\{s_{\mathrm{spec}},s_{\mathrm{coarse}}\}]\, R_{\mathrm{ret},j}, \end{equation} where $R_{\mathrm{ans}}$ evaluates the final answer obtained after executing the selected path, and $R_{\mathrm{ret}}$ is the signed retrieval gain of a generated query over the original question. For each free rollout, we derive role-specific advantages from the corresponding counterfactual rollout group: \begin{equation} A_{j,t}= \begin{cases} A_j^{\mathrm{state}}, & t\in\mathcal{T}_{\mathrm{state}}^j,\\ A_j^{\mathrm{query}}, & t\in\mathcal{T}_{\mathrm{query}}^j,\\ A_j^{\mathrm{answer}}, & t\in\mathcal{T}_{\mathrm{answer}}^j,\\ 0, & \text{otherwise}. \end{cases} \end{equation} Here, $A_j^{\mathrm{state}}$ compares the end-to-end utility of the four access paths, $A_j^{\mathrm{query}}$ compares generated queries within the same retrieval state, and $A_j^{\mathrm{answer}}$ optimizes the direct-answer span. $\mathcal{T}_{\mathrm{state}}^j$, $\mathcal{T}_{\mathrm{query}}^j$, and
$\mathcal{T}_{\mathrm{answer}}^j$ denote the state, query, and answer spans in
rollout $j$, respectively. We incorporate these token-level advantages into a DAPO-style policy objective~\citep{yu2025dapo}. Complete reward and advantage are provided in \hyperlink{supplement.1}{the supplementary material}.

\section{Experiments}

We evaluate GistPilot around four research questions:
\begin{itemize}
\item \textbf{RQ1:} Does GistPilot improve end-to-end answer quality and evidence retrieval?
\item \textbf{RQ2:} What are the contributions of each component?
\item \textbf{RQ3:} Does GAP select appropriate execution paths rather than relying on a fixed strategy?
\item \textbf{RQ4:} How efficient is GistPilot, and how does UGT capacity affect its performance?
\end{itemize}

\subsection{Experimental Setup}
\begin{table*}[t]
\centering
\setlength{\tabcolsep}{2.8pt}
\resizebox{\textwidth}{!}{%
\begin{tabular}{@{}l*{18}{c}@{}}
\toprule
\textbf{Method}
& \multicolumn{3}{c}{\textbf{RAG}}
& \multicolumn{3}{c}{\textbf{MILL}}
& \multicolumn{3}{c}{\textbf{HyDE}}
& \multicolumn{3}{c}{\textbf{RF-Mem}}
& \multicolumn{3}{c}{\textbf{PBR}}
& \multicolumn{3}{c}{\textbf{Ours}} \\
\textbf{Metrics}
& \textbf{Judge} & \textbf{R@5} & \textbf{N@5}
& \textbf{Judge} & \textbf{R@5} & \textbf{N@5}
& \textbf{Judge} & \textbf{R@5} & \textbf{N@5}
& \textbf{Judge} & \textbf{R@5} & \textbf{N@5}
& \textbf{Judge} & \textbf{R@5} & \textbf{N@5}
& \textbf{Judge} & \textbf{R@5} & \textbf{N@5} \\
\cmidrule(lr){1-1}
\cmidrule(lr){2-4}
\cmidrule(lr){5-7}
\cmidrule(lr){8-10}
\cmidrule(lr){11-13}
\cmidrule(lr){14-16}
\cmidrule(lr){17-19}

\rowcolor{retrievergray}
\multicolumn{19}{c}{\textit{bge-m3}} \\

\textbf{Overall}
& 0.285 & 0.418 & 0.345
& 0.272 & 0.359 & 0.318
& 0.249 & 0.387 & 0.327
& \underline{0.294} & 0.418 & 0.337
& 0.275 & \underline{0.434} & \underline{0.396}
& \textbf{0.314} & \textbf{0.464} & \textbf{0.398} \\

Basic information
& \underline{0.282} & \underline{0.471} & 0.324
& \underline{0.282} & 0.402 & 0.300
& 0.227 & 0.405 & 0.281
& 0.273 & 0.453 & 0.300
& 0.255 & 0.470 & \underline{0.369}
& \textbf{0.327} & \textbf{0.539} & \textbf{0.400} \\

Preference (easy)
& 0.423 & 0.433 & 0.397
& 0.327 & 0.314 & 0.346
& \underline{0.442} & 0.436 & 0.427
& 0.397 & 0.417 & 0.374
& 0.429 & \underline{0.472} & \underline{0.446}
& \textbf{0.455} & \textbf{0.551} & \textbf{0.468} \\

Preference (hard)
& \underline{0.307} & 0.368 & 0.392
& 0.268 & 0.343 & 0.394
& 0.288 & 0.353 & \underline{0.425}
& 0.302 & \underline{0.413} & 0.422
& \textbf{0.317} & \textbf{0.415} & \textbf{0.462}
& 0.278 & 0.374 & 0.407 \\

Social
& 0.208 & 0.339 & 0.325
& 0.226 & 0.305 & 0.280
& 0.170 & 0.353 & 0.297
& \textbf{0.283} & 0.349 & 0.330
& 0.208 & \underline{0.355} & \textbf{0.376}
& \underline{0.245} & \textbf{0.378} & \underline{0.344} \\

\midrule
\rowcolor{retrievergray}
\multicolumn{19}{c}{\textit{all-MiniLM-L6-v2 }} \\

\textbf{Overall}
& \underline{0.276} & 0.367 & 0.313
& 0.217 & 0.326 & 0.278
& 0.248 & \underline{0.381} & \underline{0.329}
& 0.237 & 0.278 & 0.243
& 0.238 & 0.349 & 0.296
& \textbf{0.304} & \textbf{0.408} & \textbf{0.351} \\

Basic information
& \underline{0.273} & 0.352 & \underline{0.264}
& 0.200 & 0.320 & 0.230
& 0.236 & \underline{0.402} & \textbf{0.294}
& 0.255 & 0.273 & 0.193
& 0.245 & 0.383 & 0.256
& \textbf{0.309} & \textbf{0.403} & \textbf{0.294} \\

Preference (easy)
& \underline{0.429} & 0.421 & 0.417
& 0.401 & \underline{0.437} & \textbf{0.449}
& \textbf{0.465} & 0.424 & 0.425
& 0.327 & 0.310 & 0.313
& 0.330 & 0.335 & 0.356
& 0.413 & \textbf{0.464} & \underline{0.443} \\

Preference (hard)
& \textbf{0.298} & \underline{0.341} & 0.369
& 0.205 & 0.230 & 0.278
& \underline{0.268} & \textbf{0.364} & \textbf{0.403}
& \underline{0.268} & 0.301 & 0.330
& \textbf{0.298} & 0.298 & 0.363
& \textbf{0.298} & \textbf{0.364} & \underline{0.392} \\

Social
& \underline{0.189} & \underline{0.393} & \underline{0.319}
& 0.170 & 0.361 & 0.292
& 0.151 & 0.328 & 0.296
& 0.132 & 0.256 & 0.246
& 0.132 & 0.326 & 0.296
& \textbf{0.245} & \textbf{0.428} & \textbf{0.353} \\

\midrule
\rowcolor{retrievergray}
\multicolumn{19}{c}{\textit{all-mpnet-base-v2  }} \\

\textbf{Overall}
& 0.266 & 0.377 & 0.315
& 0.247 & 0.313 & 0.274
& \underline{0.280} & 0.391 & 0.311
& 0.240 & 0.337 & 0.262
& 0.245 & \textbf{0.427} & \textbf{0.366}
& \textbf{0.292} & \underline{0.394} & \underline{0.348} \\

Basic information
& 0.273 & 0.424 & 0.310
& 0.291 & 0.338 & 0.260
& \underline{0.300} & 0.433 & 0.304
& 0.236 & 0.365 & 0.254
& 0.264 & \textbf{0.486} & \textbf{0.377}
& \textbf{0.318} & \underline{0.447} & \underline{0.349} \\

Preference (easy)
& \underline{0.362} & 0.429 & 0.411
& 0.308 & 0.369 & 0.355
& \textbf{0.404} & \textbf{0.485} & \underline{0.432}
& \underline{0.362} & 0.385 & 0.341
& \underline{0.362} & \underline{0.446} & \textbf{0.433}
& 0.346 & 0.441 & 0.417 \\

Preference (hard)
& 0.288 & 0.366 & \underline{0.414}
& 0.288 & 0.340 & 0.367
& 0.288 & 0.370 & 0.394
& 0.288 & 0.329 & 0.337
& \underline{0.293} & \underline{0.374} & 0.388
& \textbf{0.322} & \textbf{0.385} & \textbf{0.428} \\

Social
& \textbf{0.189} & 0.263 & 0.202
& 0.094 & 0.213 & 0.191
& \underline{0.170} & 0.275 & 0.202
& 0.151 & 0.261 & 0.184
& 0.113 & \textbf{0.336} & \textbf{0.294}
& \textbf{0.189} & \underline{0.285} & \underline{0.227} \\

\midrule
\textbf{Overall Average}
& \underline{0.276} & 0.387 & 0.324
& 0.245 & 0.333 & 0.290
& 0.259 & 0.386 & 0.322
& 0.257 & 0.344 & 0.281
& 0.253 & \underline{0.403} & \underline{0.353}
& \textbf{0.303} & \textbf{0.422} & \textbf{0.366} \\

\bottomrule
\end{tabular}%
}
\caption{End-to-end answer quality and retrieval performance on PersonaBench across three dense retrievers. Bold and underline indicate the best and second-best results.}
\label{tab:Personabench_main}
\end{table*}
\paragraph{Datasets.}
Following the construction protocol of LongMemEval~\citep{wu2025longmemeval},
we build roughly $4{,}000$ samples to train the access policy, and evaluate on
two personalized long-term memory benchmarks. \textbf{LongMemEval}, for which
we use the LongMemEval-S variant and refer to as LongMemEval hereafter,
contains 500 questions spanning seven abilities: temporal reasoning,
multi-session reasoning, knowledge update, user-side and assistant-side
single-session information, single-session preference, and abstention. \textbf{PersonaBench}~\citep{tan2025personabench} contains 230
questions that probe a model's understanding of a user's basic attributes,
preferences, and social relations. Full construction and dataset details are
given in \hyperlink{supplement.1}{the supplementary material}.
\begin{figure}[t]
\centering
\includegraphics[width=\columnwidth]{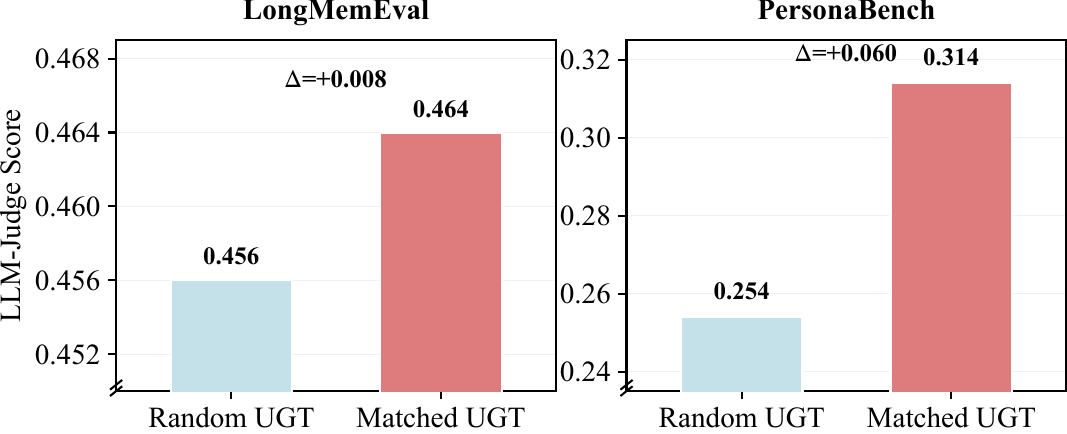}
\caption{LLM-judge scores under a matched versus a randomly selected user's
UGT on LongMemEval and PersonaBench. A matched UGT consistently outperforms a
random one, with the gap especially pronounced on PersonaBench.}
\label{fig:kv-match}
\end{figure}
\paragraph{Metrics.}
We report both end-to-end answer quality and evidence retrieval quality. For
answer quality, a \textbf{LLM judge} decides whether a predicted answer
satisfies the facts required by the reference, and \textbf{word-level F1}
measures token overlap with the reference. For LongMemEval, we use its official LLM-judge protocol; for PersonaBench, we construct a judge following the same protocol. For retrieval quality,
\textbf{Recall@5} measures coverage of the target evidence within the top five
results and \textbf{nDCG@5} measures its ranking quality.
\begin{table}[t]
\centering
\setlength{\tabcolsep}{3pt}
\resizebox{\columnwidth}{!}{%
\begin{tabular}{lcccccccc}
\toprule
Method & Temp. & MS & KU & SS-U & SS-A & SS-P & AB & Overall \\
\midrule
\textbf{GistPilot} & \underline{0.263} & \textbf{0.376} & \textbf{0.564} & \textbf{0.743} & 0.839 & \underline{0.133} & \textbf{0.633} & \textbf{0.464} \\
RAG    & \underline{0.263} & \underline{0.338} & \underline{0.549} & \underline{0.729} & \underline{0.893} & \textbf{0.167} & \underline{0.567} & \underline{0.458} \\
HyDE   & 0.256 & 0.263 & 0.462 & 0.700 & 0.821 & \textbf{0.167} & 0.400 & 0.410 \\
RF-Mem & 0.233 & 0.331 & 0.513 & 0.614 & \textbf{0.911} & \textbf{0.167} & 0.333 & 0.428 \\
MILL   & \textbf{0.286} & 0.278 & 0.500 & 0.686 & 0.821 & \underline{0.133} & 0.300 & 0.424 \\
PBR    & 0.233 & 0.331 & 0.436 & 0.700 & 0.821 & \textbf{0.167} & 0.367 & 0.418 \\
Fullcontext & 0.143 & 0.090 & 0.128 & 0.243 & 0.232 & \textbf{0.167} & 0.033 & 0.152 \\
Base   & 0.165 & 0.053 & 0.077 & 0.086 & 0.143 & 0.000 & 0.133 & 0.098 \\
\bottomrule
\end{tabular}%
}
\caption{End-to-end answer quality by question type on LongMemEval. Temp., MS, KU, SS-U, SS-A, SS-P, and AB denote temporal reasoning, multi-session reasoning, knowledge update, single-session user information, single-session assistant information, single-session preference, and abstention, respectively.}
\label{tab:type-lme}
\end{table}
\paragraph{Baselines.}
We compare GistPilot against seven baselines that cover the main ways of using
long-term user history. \textbf{Base} answers from the model alone, with no
external memory. \textbf{Full-context} concatenates as much user history as
the context window allows and answers without retrieval. \textbf{RAG}
\citep{lewis2020rag} retrieves with the original question and reads the top
passages. \textbf{HyDE}~\citep{gao2023hyde}
generates a hypothetical answer document and retrieves with its embedding.
\textbf{MILL}~\citep{jia2024mill} generates candidate expansion documents and
selects them by mutual verification against pseudo-relevance evidence.
\textbf{PBR}~\citep{zhang2026pbr} rewrites the query with personalized signals
derived from the user profile before retrieval. \textbf{RF-Mem}
\citep{zhang2026rfmem} estimates familiarity from retrieval scores and switches
between a one-shot and an iterative recollection path.
\begin{figure*}[t]
\centering
\includegraphics[width=\textwidth]{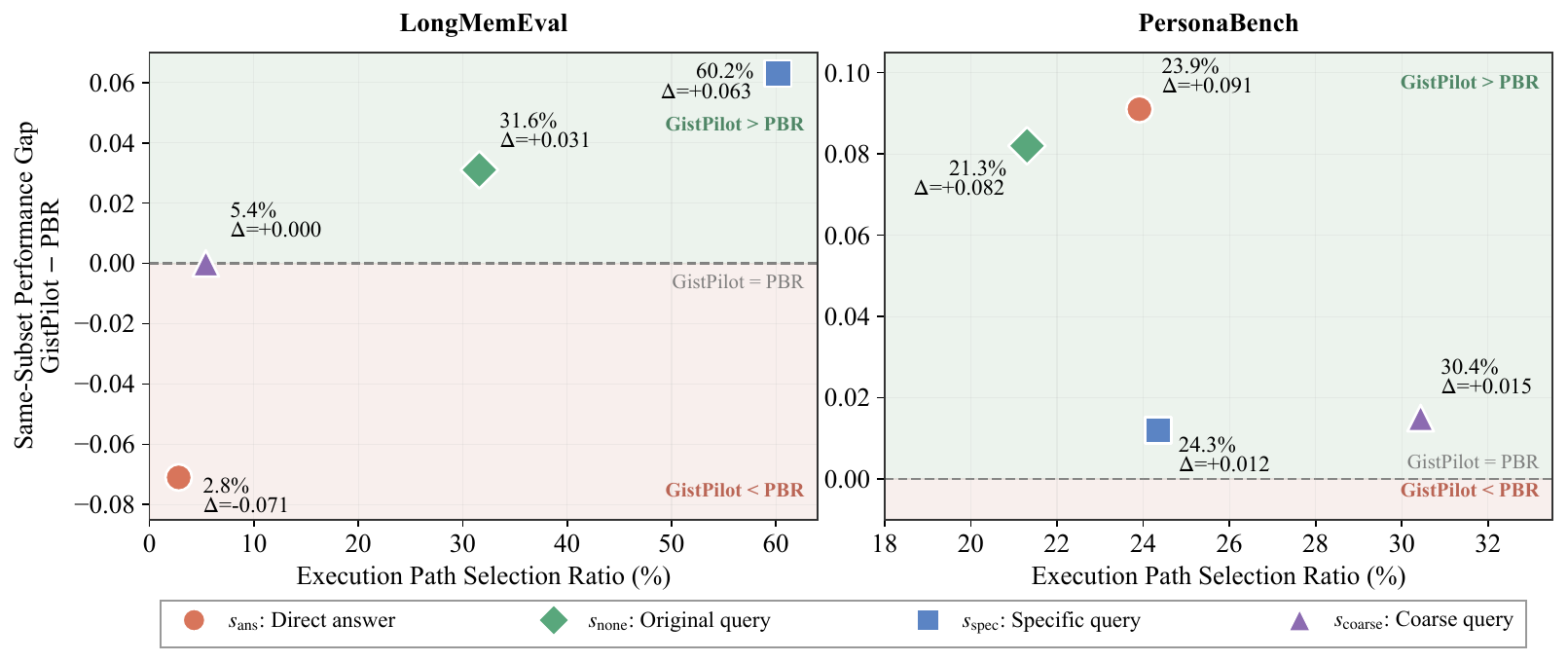}
\caption{GAP path selection frequency and same-subset performance advantage.
For each execution path, the horizontal axis reports how often GAP selects the
path and the vertical axis reports the performance gap between GistPilot and PBR
on the same subset of questions (positive means GistPilot is better). PersonaBench results in this figure use all-mpnet-base-v2.}
\label{fig:path}
\end{figure*}
\paragraph{Implementation.}
We train GistPilot on top of Qwen2.5-7B-Instruct~\citep{qwen2024qwen25} and use
a separate, untrained Qwen2.5-7B-Instruct as the answerer. On PersonaBench we
evaluate across three dense retrievers, BGE-M3~\citep{chen2024bgem3},
all-MiniLM-L6-v2~\citep{wang2020minilm}, and
all-mpnet-base-v2~\citep{song2020mpnet}, to assess robustness to the choice of
retriever; on LongMemEval we use BGE-M3 as the single
retriever. In all cases we retrieve the top five passages and cap retrieval
context at 16k tokens. All experiments are conducted using eight NVIDIA H20
GPUs.

\subsection{Overall Performance (RQ1)}
\paragraph{GistPilot consistently delivers the best end-to-end answer quality.}
Tables~\ref{tab:Personabench_main} and ~\ref{tab:type-lme} report the
PersonaBench and LongMemEval results, respectively. Across both benchmarks, GistPilot achieves the highest overall Judge score,
reaching 0.464 on LongMemEval and 0.314 with BGE-M3 on PersonaBench. Across all
three PersonaBench retrievers, it consistently ranks first in overall Judge and
remains first or second on both Recall@5 and nDCG@5, demonstrating robustness
to the choice of retriever. Overall, these results demonstrate the effectiveness
of the GistPilot framework for personalized long-term-memory question
answering. Results for the Base and Full-context baselines on PersonaBench, along with word-level F1 scores, are reported in \hyperlink{supplement.1}{the supplementary material}.

\begin{table}[t]
\centering
\setlength{\tabcolsep}{4pt}
\resizebox{\columnwidth}{!}{%
\begin{tabular}{lccccc}
\toprule
Variant & UGT & SFT & RL & LME & PB \\
\midrule
\textbf{GistPilot} & $\bullet$ & $\bullet$ & $\bullet$
    & \textbf{0.464} & \textbf{0.314} \\
w/o RL             & $\bullet$ & $\bullet$ & --
    & 0.394 & 0.267 \\
w/o SFT            & $\bullet$ & -- & $\bullet$
    & \underline{0.446} & 0.270 \\
w/o SFT and RL     & $\bullet$ & -- & --
    & 0.164 & 0.199 \\
w/o UGT            & -- & $\bullet$ & $\bullet$
    & 0.443 & 0.272 \\
w/o forced rollout & $\bullet$ & $\bullet$ & $\bullet$
    & 0.438 & \underline{0.297} \\
\bottomrule
\end{tabular}
}
\caption{Ablation of the UGT and GAP optimization components.}
\label{tab:stage}
\end{table}

\paragraph{GistPilot is particularly effective when answering requires a holistic view of user history.}The category-level results in Tables~\ref{tab:Personabench_main} and
\ref{tab:type-lme} show clear advantages on multi-session reasoning,
knowledge update, and abstention, while remaining competitive on temporal
reasoning, with particularly notable gains of 0.038 and 0.066 over fixed RAG
on multi-session and abstention questions. On PersonaBench, GistPilot also
remains among the strongest methods across all four question types.
These tasks require integrating evidence across sessions, tracking changes over time, or recognizing when the history is insufficient. Similarity-based retrieval may identify a locally
relevant fragment but still miss the broader context needed for a complete
answer. UGT addresses this limitation by preserving a coarse-grained global
impression of the user's history, while GAP uses that trace to select an
appropriate answering or retrieval strategy.

\subsection{Ablation Analysis (RQ2)}

We evaluate GistPilot through two complementary ablation studies, focusing on the global guidance provided by UGT and the contributions of the remaining components.

\paragraph{The user-specific gist trace contributes information that cannot be replaced by an another user's latent trace.}
As shown in Figure~\ref{fig:kv-match}, replacing the matched UGT with a randomly
selected user's trace reduces the Judge score by 0.060 on PersonaBench and by
0.008 on LongMemEval. These results show that UGT gives the model a global view of the user's history before retrieval, allowing subsequent memory access to be guided by more than local
similarity.

\paragraph{The two-stage optimization is essential for turning the gist trace into effective access decisions.}
As shown in Table~\ref{tab:stage}, the complete model outperforms variants that
remove either SFT or RL, while omitting both stages causes a substantial decline
on both datasets. Replacing forced multi-path rollout with free rollout consistently degrades performance. These trends show that SFT first establishes
stable retention-state boundaries and conditional generation behavior, whereas
outcome-driven RL refines path selection and generated content according to
final answer quality. The benefit of forced rollout further confirms
that comparing alternative access paths on the same question is important for
learning when and how GAP should use external memory.

\subsection{Adaptive Path Selection (RQ3)}

We assess GAP from two complementary perspectives: whether adaptive routing
outperforms forcing a single path for every question, and how the learned
policy distributes questions across the four execution paths. Results are
reported in Table~\ref{tab:fixed} and Figure~\ref{fig:path}.

\begin{table}[t]
\centering
\setlength{\tabcolsep}{5.5pt}
\renewcommand{\arraystretch}{1.08}
\begin{tabular}{@{}lc@{}}
\toprule
Access strategy & Judge $\uparrow$ \\
\midrule
Direct answering ($s_{\mathrm{ans}}$)                 & 0.218 \\
Original-query retrieval ($s_{\mathrm{none}}$)       & \underline{0.285} \\
Specific-query retrieval ($s_{\mathrm{spec}}$)       & 0.283 \\
Coarse-query retrieval ($s_{\mathrm{coarse}}$)       & 0.259 \\
Adaptive routing (GistPilot)                          & \textbf{0.314} \\
\bottomrule
\end{tabular}
\caption{Adaptive routing versus fixed access strategies on PersonaBench.}
\label{tab:fixed}
\end{table}

\paragraph{Adaptive routing is more effective than committing to any single execution path.}
As shown in Table~\ref{tab:fixed}, GistPilot reaches a Judge score of 0.314 on
PersonaBench, compared with 0.285 for the strongest fixed strategy. The same
path is therefore not suitable for every question: some can be answered from
the retained trace, whereas others require different levels of external memory
access. The advantage of adaptive routing confirms that GAP can turn the
memory signal in UGT into an appropriate execution decision rather than
applying a uniform retrieval strategy.

\paragraph{The four execution paths are complementary, and GAP suppresses paths that are less reliable.}Figure~\ref{fig:path} compares each selected path with PBR on exactly the same
subset of questions, thereby separating path utility from differences in
subset difficulty. On PersonaBench, all four paths outperform PBR, with the
largest gains from direct answering and original-query retrieval, while their
balanced selection frequencies show that each path serves a meaningful subset
of questions. LongMemEval exhibits stronger specialization: precise- and
original-query retrieval jointly cover 91.8\% of the questions and both improve
over PBR, whereas direct answering underperforms and is selected for only
2.8\%; coarse-query retrieval is also rarely selected and matches PBR on its
subset. These results show that the four paths provide complementary options
and that GAP routes questions toward those that are reliable for the current
benchmark.

\subsection{Efficiency and UGT Capacity (RQ4)}

Finally, we examine the practical trade-offs of GistPilot in terms of
end-to-end inference latency and the capacity of its fixed-length user trace,
as reported in Tables~\ref{tab:latency} and~\ref{tab:kvlen}.

\begin{table}[t]
\centering
\setlength{\tabcolsep}{3pt}
\renewcommand{\arraystretch}{1.06}
\resizebox{\columnwidth}{!}{%
\begin{tabular}{@{}lccccccc@{}}
\toprule
Method
& Full
& RAG
& HyDE
& MILL
& PBR
& RF-Mem
& Ours \\
\midrule
LME
& 40.1
& \textbf{26.7}
& 53.3
& 36.5
& 37.8
& 27.8
& \underline{27.6} \\
PB
& 43.0
& \underline{22.8}
& 43.0
& 30.5
& 32.8
& 23.3
& \textbf{14.6} \\
\bottomrule
\end{tabular}
}
\caption{Average end-to-end inference latency in seconds per question.}
\label{tab:latency}
\end{table}

\begin{table}[t]
\centering
\setlength{\tabcolsep}{7pt}
\renewcommand{\arraystretch}{1.06}
\resizebox{\columnwidth}{!}{%
\begin{tabular}{@{}lccccc@{}}
\toprule
UGT length
& 64
& 128
& 256
& 512
& 1024 \\
\midrule
Judge
& 0.125
& 0.140
& \textbf{0.164}
& \underline{0.145}
& 0.130 \\
\bottomrule
\end{tabular}
}
\caption{Effect of UGT prefix length on direct answering from UGT alone on LongMemEval.}
\label{tab:kvlen}
\end{table}

\paragraph{GistPilot improves answer quality without the latency overhead of multi-stage query expansion.}
Table~\ref{tab:latency} shows that GistPilot is the fastest method on
PersonaBench at 14.6 seconds per question and remains close to standard RAG on
LongMemEval. It is consistently more efficient than methods that generate or
verify additional retrieval content, including HyDE, MILL, and PBR. This trend
shows that adaptive execution does not merely improve accuracy: by answering
directly when possible and invoking only the required retrieval path, GistPilot
also avoids unnecessary inference cost.

\paragraph{UGT benefits from sufficient but not excessive capacity.}
In the direct-answer setting used to isolate trace capacity,
Table~\ref{tab:kvlen} shows that performance improves as the KV length grows
from 64 to 256, where the Judge score peaks at 0.164, but declines with longer
prefixes. A short trace cannot retain enough long-term information, whereas an
overly large trace introduces additional optimization difficulty without a
corresponding gain. These results indicate that a moderate UGT capacity
provides the best trade-off between representational capacity and optimization
stability, with a KV length of 256 achieving the strongest performance.

\section{Conclusion}
We presented GistPilot, a personalized memory framework that couples a compact
internal user trace with adaptive access to external history. UGT compresses
long-term user interactions into a fixed-length latent KV memory that provides
a holistic view of the user before retrieval, while GAP estimates the
question-relevant memory retained in this trace and selects among direct
answering and three retrieval strategies. A supervised warm start followed by
outcome-driven reinforcement learning enables the policy to learn both access
decisions and state-conditioned content from downstream utility. Experiments
on LongMemEval and PersonaBench show that GistPilot consistently improves
end-to-end answer quality, remains robust across retrievers, and avoids the
overhead of uniformly applying complex retrieval pipelines. Ablation and
routing analyses further confirm the value of user-specific gist traces,
outcome-driven optimization, and adaptive path selection. These results
establish internal memory awareness as an effective basis for more accurate
and efficient access to long-term user history.
\bibliographystyle{plainnat}
\bibliography{gistpilot}

\clearpage
\includepdf[pages=-,fitpaper=true,link,linkname=supplement]{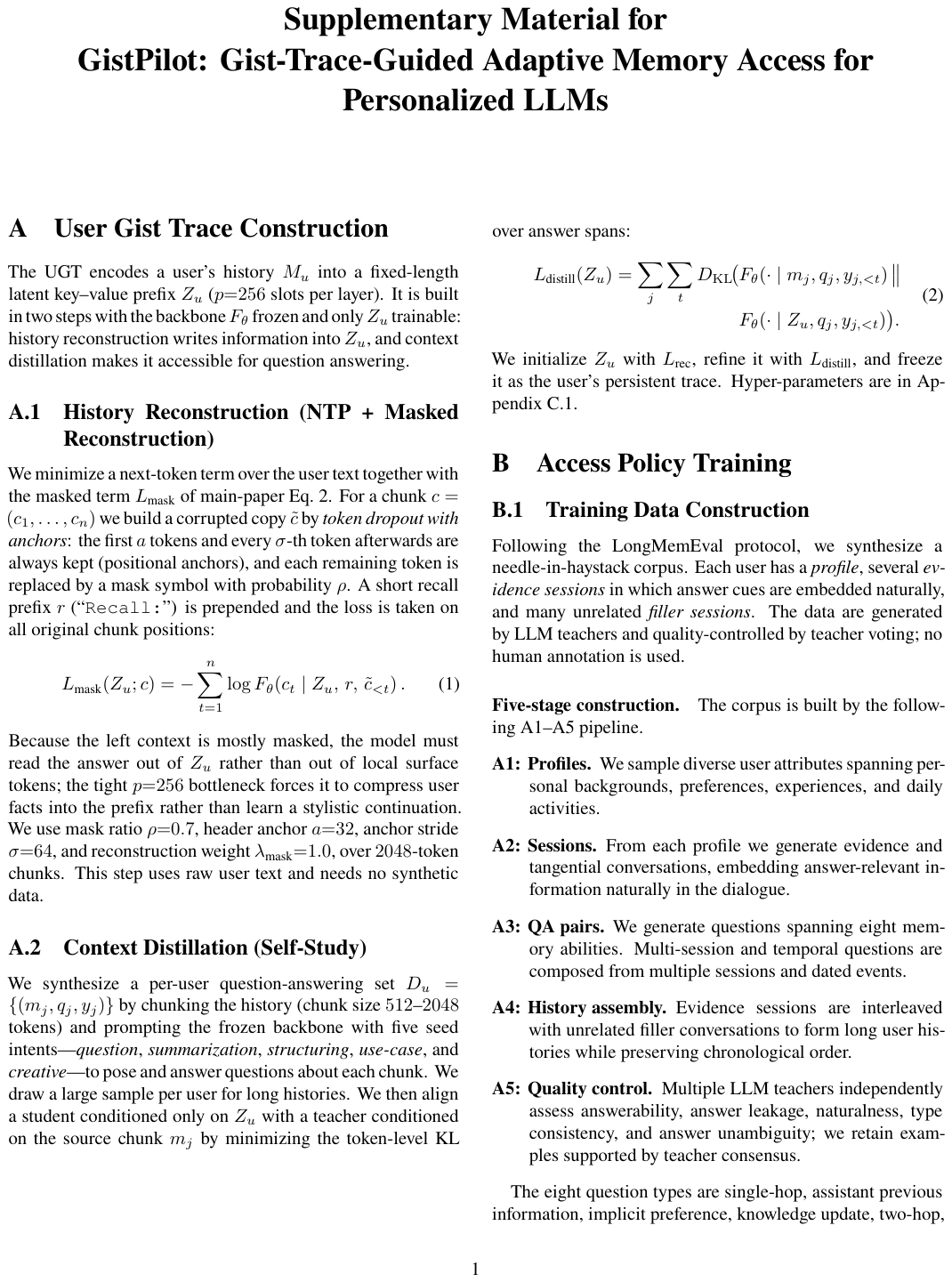}

\end{document}